\documentclass{article}
\usepackage{spconf,amsmath,graphicx}

\usepackage{algorithm}
\usepackage{algorithmic}

\usepackage{multirow} 
\usepackage{multicol}  

\title{
	A novel centroid update approach for clustering-based superpixel methods and superpixel-based edge detection 
}

\name{Houwang Zhang$^{\star}$ \qquad Chong Wu$^{\dagger}$ \qquad Le Zhang$^{\ddagger}$ \qquad Hanying Zheng$^{\star}$}

\address{$^{\star}$ School of Automation, China University of Geosciences, Wuhan, China \\
	$^{\dagger}$ Department of Electrical Engineering, City University of Hong Kong, Kowloon, Hong Kong\\
	$^{\ddagger}$ Department of Computer Science and Technology, Tongji University, Shanghai, China}

\begin{document}

	\maketitle
	
	\begin{abstract}
	Superpixel is widely used in image processing. And among the methods for superpixel generation, clustering-based methods have a high speed and a good performance at the same time. However, most clustering-based superpixel methods are sensitive to noise. To solve these problems, in this paper, we first analyze the features of noise. Then according to the statistical features of noise, we propose a novel centroid update approach to enhance the robustness of clustering-based superpixel methods. Besides, we propose a novel superpixel-based edge detection method. The experiments on BSD500 dataset show that our approach can significantly enhance the performance of clustering-based superpixel methods in noisy environment. Moreover, we also show that our proposed edge detection method outperforms other classical methods.
	\end{abstract}

	
	\section{Introduction}
	\label{sec:intro}
	
	In 2003, Ren \emph{et al}. \cite{ren2003learning} first proposed the concept of superpixel, which refers to a series of regions consists of pixels with adjacent positions and similar color, brightness and texture features. These regions can retain the effective information like the boundary information of objects in the image \cite{fslic, ICIP, accw}. Different from pixel, superpixel can reduce the size of the object to be processed and the complexity of the subsequent processing to a great extent \cite{fslic, ICIP, accw}. Due to these advantages, superpixel methods are usually used as a preprocessing step for many tasks \cite{fslic, ICIP, accw, achanta2017superpixels}.
	
	For the past years, many superpixel methods have been proposed. A good superpixel method should meet many needs like compactness, boundary adherence, computational complexity and controllable superpixel number \cite{fslic, achanta2012slic}. Each kind of superpixel method has its own advantages and defects \cite{fslic, jampani2018superpixel}. Among them, clustering-based methods are widely used for image segmentation tasks \cite{fslic, guo2018fuzzy}. Through the clustering process, the number and the compactness of superpixels can be controlled \cite{stutz2018superpixels}. Simple linear iterative clustering (SLIC) is one of the most commonly used clustering-based methods, which adopts a local K-means clustering method to cluster pixels based on the color and spatial distance \cite{achanta2012slic}. Linear spectral clustering (LSC) is another well-known clustering-based method \cite{li2015superpixel}. Different to the five-dimensional space used in SLIC, it takes a ten-dimensional space and gets a better boundary recall rate than SLIC. Recently, an improved SLIC called simple non-iterative clustering (SNIC) has been developed \cite{achanta2017superpixels}. Compared to SLIC and LSC, SNIC do not need iterations, thus it has higher computational and memory efficiency. However, above clustering-based methods are all sensitive to noise \cite{fslic}. When the noise exists, they can not maintain the performance as they work in noise-free situation \cite{fslic}.
	
	To solve the above problems of clustering-based superpixel methods, in this paper, we first analyze the features of noise. Then according to the statistical features of noise, we propose a novel centroid update approach to enhance clustering-based superpixel methods. Moreover, we propose a superpixel-based edge detection algorithm (SBED), which can gain the edge of image by detecting edges of superpixels. The contribution of this paper can be concluded as follows,
	
	\begin{itemize}
		\setlength{\itemsep}{0pt}
		\setlength{\parsep}{0pt}
		\setlength{\parskip}{0pt}
		\item We analyze the reason why clustering-based superpixel methods don't work well in noisy environment.
		
		\item According to the features of noise, we propose a novel centroid update approach for clustering-based superpixel methods to reduce the impact of noise.
		
		\item Based on superpixel segmentation, we propose a new edge detection method.
	\end{itemize}

	\section{Methods}
	\label{sec:format}
	
	\subsection{A novel centroid update approach}
	
	When noise exists, value of pixel tends to be singular. While the cluster centroid usually takes the mean value of all pixels with corresponding label, the impact of noise will be accumulated in the cluster centroid \cite{kslic}. As we know, the cluster centroid plays an important role in the process of clustering.
	
	\begin{figure}[!htbp]
		\centerline{\includegraphics[width = 1\columnwidth]{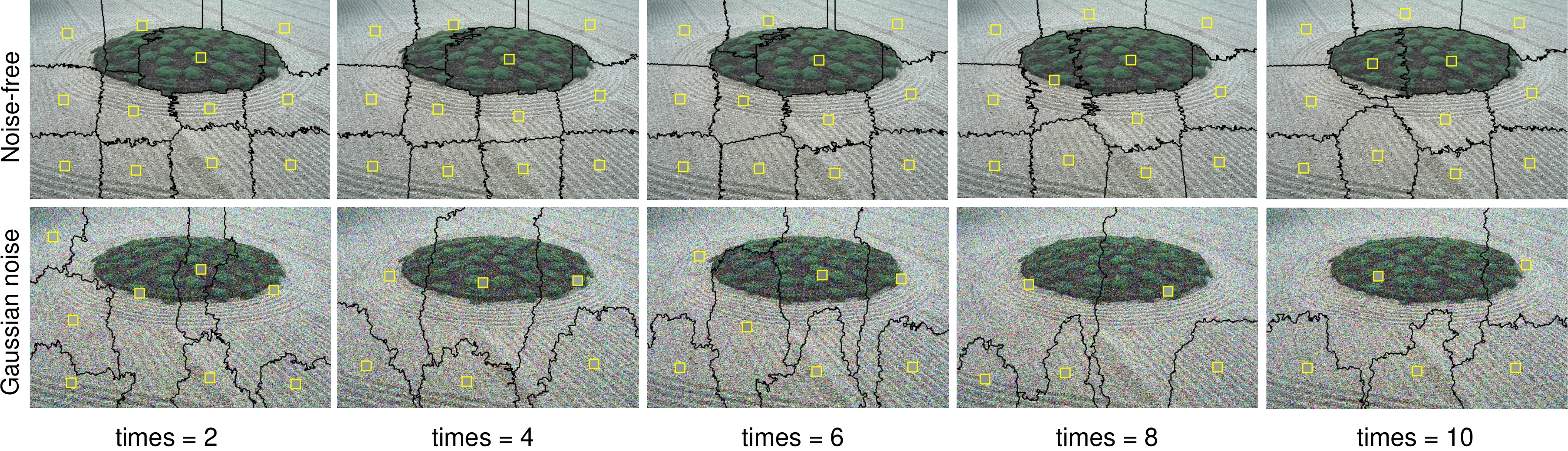}}
		\caption{Performance of SLIC in noise-free environment and Gaussian noise environment with different iteration times.}
		\label{fig1}
	\end{figure}
	
	Take SLIC for example, as shown in the Fig. \ref{fig1}, in the early stage of iterative clustering in noise-free environment, the segmentation result has some false boundaries. Generally, the error will reduce and converge with the increase of iteration times. But in noisy environment, the existence of noise will lead to a worse false segmentation, which will lead to more errors on the cluster centroids as a positive feedback. As Fig. \ref{fig1} shows, in noisy environment, the difference of cluster centroids becomes fuzzy, and cluster centroids can't capture the features of the expected object but the mixed region caused by false segmentation. Finally, it will results in a poor segmentation.
	
	Distribution of most noise follows or approximates to the Gaussian distribution as follows:
	\begin{equation}
	Noise_{(z)} = \frac{1}{\sqrt{2\pi}\delta }exp[\frac{-(z - u)^{2}}{2\delta ^{2}}],
	\label{eq1}
	\end{equation}
	where $z$ is the value of the pixel, $u$ denotes the average or expected value of $z$, $\delta$ denotes the standard deviation of $z$.
	
	As Eq. (\ref{eq1}) shows, the value of $Noise_{(z)}$ is distributed on both sides of $u$. Hence, most denoising methods take the neighbors of the pixel to eliminate the impact of noise, such as the works in \cite{buades2005non, zhang2014new}, they deal with the noisy pixel by taking the mean value of pixels within a square window centered at the current pixel.
	
	Considering the statistical features of noise and inspired by these denoising methods, we propose a novel centroid update approach and apply it to clustering-based superpixel methods. For an image with size $M \times N$ and desired number of superpixels $k$, $Centroid_i$ of the $i$th superpixel can be computed as follows:
	

\begin{equation}
\begin{split}
Centroid_i = \mathbf{[C_{i,a}, C_{i,b}] }^\mathrm{T},\\
C_{i,a} = [C_{i,1}, \cdots , C_{i,m}], C_{i,m} =  \frac{\sum_{p_{x, y} \in S_{i}} q_{x, y}^{m}}{\left | S_i \right |},\  \\
C_{i,b} = [C_{i,m + 1}, \cdots , C_{i,n}], C_{i,n} = \frac{\sum_{p_{x, y} \in B_{i}}q_{x, y}^{n}}{\left | B_i \right |},\  
\end{split}
\label{eq2}
\end{equation}

	where $C_{i, a}$ represents the spatial part (spatial centroid) of $Centroid_i$, $C_{i, b}$ is the color part (color centroid), and $S_i$ represents the region of the $i$th superpixel, $\left | S_i \right |$ is the number of pixels within $S_i$, $q_{x, y}^{m}$ is the value of pixel $p_{x, y}$ at current coordinate $m$. Because the spatial space is not affected by noise, here we still use all the pixels within the current superpixel to calculate $C_{i, a}$. Different to the spatial space, the color space will be strongly affected by noise. Thus when calculate the color centroid $C_{i, b}$ of the $i$th superpixel, instead of using all pixels in a superpixel, we use a square block $B_i$ centered at $C_{i, a}$ with an adaptive size $\sqrt{\frac{M \times  N}{k \times 2}}$ to select the pixels and calculate it, here $\left | B_i \right |$ approximates to half of $\left | S_i \right |$. There are two advantages to do like this: (1) it can reduce the effect of noise on the cluster centroids by taking the mean value of pixels within the square block; (2) it also avoids the error caused by false segmentation, most pixels within $B_i$ used to compute the color centroid can capture the features of the expected object instead of the mixed region caused by false segmentation.	
	

The whole procedure of our proposed approach is presented in Algorithm \ref{alg:A}. Here for convenience we call our approach Centroid-\emph{X}, $X$ indicates a specific clustering-based superpixel method, for example, Centroid-SLIC means that the SLIC method is enhanced by our Centroid-\emph{X} approach.

	\begin{algorithm}
		\caption{Centroid-\emph{X}}
		\label{alg:A}
		\begin{algorithmic}
			\REQUIRE The superpixel segmentation method $X$, image $I$, number of desired superpixel $k$,  compactness coefficient $h$.
			\ENSURE Label matrix $L$
			
			\STATE {Initialize $k$ clustering centroids as $X$ defaults;}
			
			\WHILE {$error \geq  threshold$}
			\STATE {Segment $I$ into $k$ superpixels using the $k$ centroids;}
			\STATE {Compute label matrix $L$ as $X$ defaults;}
			\STATE {{\bfseries for} each superpixel $S_i$ {\bfseries do}}
			\STATE {\qquad Compute its spatial centroid \scalebox{0.8}{$\mathbf{C_{i,b}}$}};
			\STATE {\qquad Get the block $B_i$ centered at \scalebox{0.8}{$\mathbf{C_{i,b}}$}};
			\STATE {\qquad Based on $B_i$, get the color centroid \scalebox{0.8}{$\mathbf{C_{i,a}}$}};
			\STATE {\qquad Set centroid of $S_i$ as \scalebox{0.8}{$\mathbf{[C_{i,a}, C_{i,b}] }^\mathrm{T}$}};
			\STATE {{\bfseries end for}}
			\ENDWHILE
			\STATE {Enforce the connectivity of $L$ as $X$ defaults};
		\end{algorithmic}
	\end{algorithm}
	
	\subsection{Our proposed superpixel-based edge detection}
	
	Superpixels can well preserve the boundary of the object, so we can detect the edge of the image based on the edges of superpixels. After detecting the edges of superpixels, we use the relationship between superpixels to determine which edges should be reserved and which should be removed.
	
	For a neighboring superpixel pair: superpixel $i$ and superpixel $j$, the distance $D_{ij}$ between them is defined as follow:
	\begin{equation}
	\centering
	\begin{split}
	& D_{i,j} = \left | \bar{L}_i - \bar{L}_j \right | + \left | \bar{A}_i - \bar{A}_j \right | + \left | \bar{B}_i - \bar{B}_j \right |,\\
	& \bar{L}_i = \sum_{c \in S_i}\frac{l_c}{\left |S_i  \right |},\  \bar{A}_i = \sum_{c \in S_i}\frac{a_c}{\left |S_i  \right |}, \ \bar{B}_i = \sum_{c \in S_i}\frac{b_c}{\left |S_i  \right |},\\
	\end{split}
	\label{eq3}
	\end{equation}
	where $l_c$, $a_c$ and $b_c$ is the value of pixel in CIELAB space.

	Then we compute the adjacent matrix $\emph{\textbf{A}}$ of superpixels ($\emph{\textbf{A}}$ is the upper triangular matrix). The $i$th row of $\emph{\textbf{A}}$ consists of the distances between the $i$th superpixel and its neighbors. And for $\emph{\textbf{A}}$, its mean value $\hat{a}$ is the mean of its non-zero elements. If $A_{i,j} < \hat{a}$, the edge between $S_i$ and $S_j$ should be removed, otherwise it should be reserved. 

	Finally, we use the gradient of edges of superpixels to further detect the edge points. The whole procedure of SBED is presented in Algorithm \ref{alg:B}.

	\begin{algorithm}
	\caption{Superpixel-based edge detection}
	\label{alg:B}
	\begin{algorithmic}
		\REQUIRE Image $\emph{\textbf{I}}$, number of superpixels $k$
		\ENSURE Edge matrix $\emph{\textbf{E}}$
		
		\STATE {Segment $\emph{\textbf{I}}$ into $k$ superpixels and then detect the edges of superpixels $\emph{\textbf{E}}$;}
		
		\STATE {Gain gradient matrix $\emph{\textbf{G}}$ of $\emph{\textbf{E}}$ as Sobel \cite{kanopoulos1988design}, set $E_{i,j} = G_{i,j}$;}
		
		\STATE {Compute $\emph{\textbf{A}}$ using Eq. 3 and gain the average $\hat{a}$ of $\emph{\textbf{A}}$;}
		
		\STATE {{\bfseries for} $i = 1$ {\bfseries to} $k$ {\bfseries do}}
		\STATE {\qquad {\bfseries for} $j = i + 1$ {\bfseries to} $k$ {\bfseries do}}
		
		\STATE {\qquad \qquad {\bfseries if} $A_{i,j} < \hat{a}$ {\bfseries then} \\ \qquad \qquad \qquad Eliminate the edge of $\emph{\textbf{E}}$ between $S_i$ and $S_j$;}
		
		\STATE {\qquad \qquad{\bfseries end if}}
		\STATE {\qquad{\bfseries end for}}
		\STATE {{\bfseries end for}}
		
		\STATE {{\bfseries for} each element $G_{i,j} \in \textbf{G}_{M \times N}$  {\bfseries do}}
		
		\STATE {{\qquad \bfseries if} $G_{i,j} < G_{low}$  {\bfseries then}} $E_{i, j} = 0$; {\bfseries end if}
		
		\STATE {{\qquad \bfseries if} $G_{i,j} > G_{high}$  {\bfseries then}} $E_{i, j} = G_{i,j}$; {\bfseries end if}
		
		\STATE {{ \bfseries end for}}

		\STATE  \textbf{return} $\emph{\textbf{E}}$;
		
	\end{algorithmic}
\end{algorithm}

\section{Experiments}

In this section, we apply Centroid-\emph{X} on three common used clustering-based superpixel methods: SLIC, LSC and SNIC  about experiments are included in supplementary material. We compare the performance of the original method and the enhanced method using our approach on the Berkeley benchmark (BSD500) \cite{arbelaez2010contour} with three kinds of environment: noise-free, Gaussian noise (zero mean with standard deviation (std) range [0.1,0.2]), and salt and pepper noise (noise density range [0.1, 0.2]). All the experiments are run on a system with Intel Core i7 2.2 GHz processor, 8 GB RAM \footnote{Data and codes for our proposed Centroid-\emph{X} and SBED are published on https://github.com/ProfHubert/Centroid.}.


\subsection{Evaluation metrics and parameter settings}

We select one standard metric for compactness and two standard metrics for boundary adherence: compactness metric (CO) \cite{schick2014evaluation}, boundary recall rate (BR) \cite{levinshtein2009turbopixels}, and under segmentation error (UE) \cite{veksler2010superpixels}. Higher BR and lower UE mean a more accurate segmentation, and higher CO means a better compactness and regularity of superpixels.

In the experiments, the parameters settings between $X$ and Centroid-\emph{X} are the same. To keep the fairness, the compactness coefficient are all setting to the same level (SLIC and SNIC take compactness coefficient 30, LSC uses ratio 0.3).

\subsection{Results and analysis}
\begin{figure}
	\centerline{\includegraphics[width = 0.76\columnwidth]{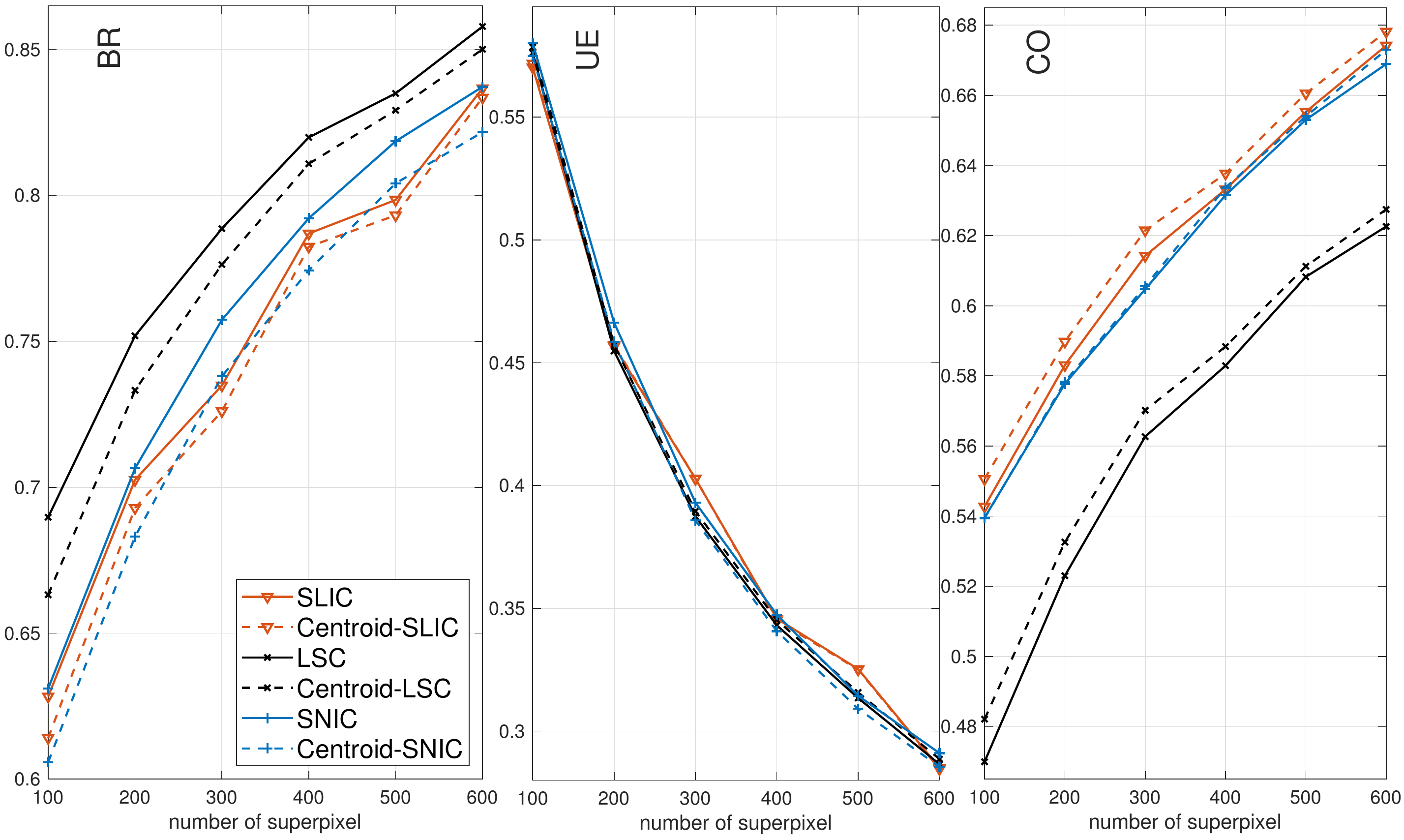}}
	\caption{The comparison of $X$ and Centroid-\emph{X} in noise-free environment.}
	\label{fig3}
\end{figure}

\begin{figure}
	\centerline{\includegraphics[width = 0.75\columnwidth]{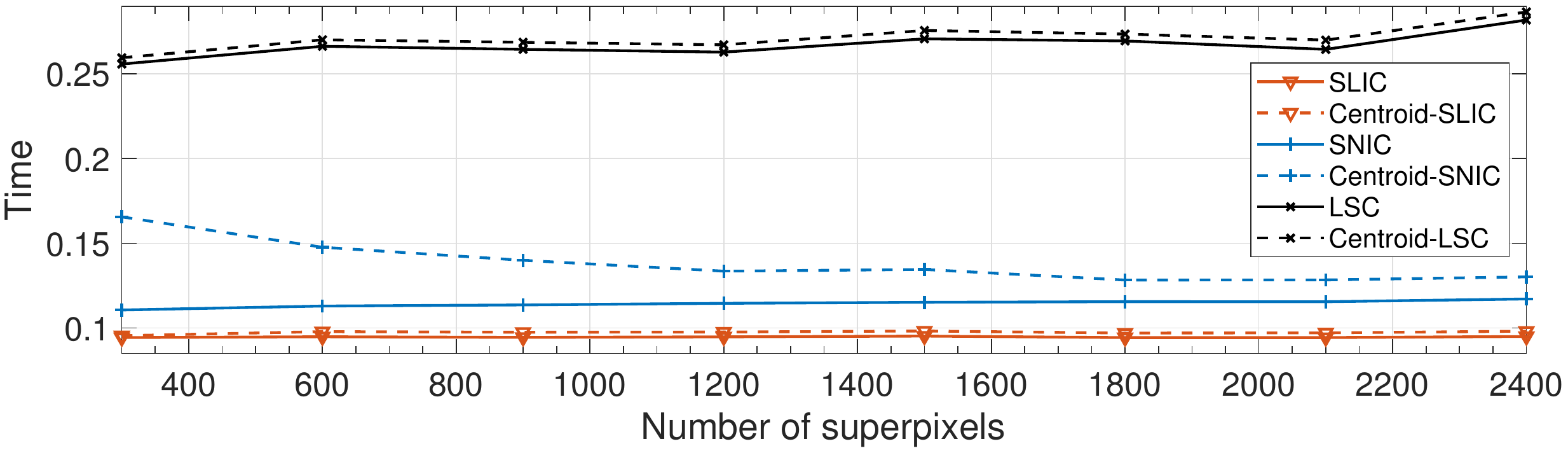}}
	\caption{The running time of $X$ and Centroid-\emph{X} in noise-free environment.}
	\label{fig4}
\end{figure}
Fig. \ref{fig3} shows the BR, UE, and CO curves of all methods in noise-free environment. By comparing them, we can find that $X$ (SLIC, LSC, SNIC) obtains better BR than Centroid-\emph{X}. But the enhanced one obtains better CO, and the UE between them is basically the same. Fig. \ref{fig4} shows their running time, except SNIC, the speed between SLIC and Centroid-SLIC and the speed between LSC and Centroid-LSC have little difference. Because SNIC takes a non-iterative clustering and its main computation focuses on the centroid update, Centroid-\emph{X} affects the computation of SNIC more than SLIC and LSC. Generally speaking, the difference between $X$ and Centroid-\emph{X} is not significant in noise-free environment.
\begin{figure}
	\centerline{\includegraphics[width = 0.8\columnwidth]{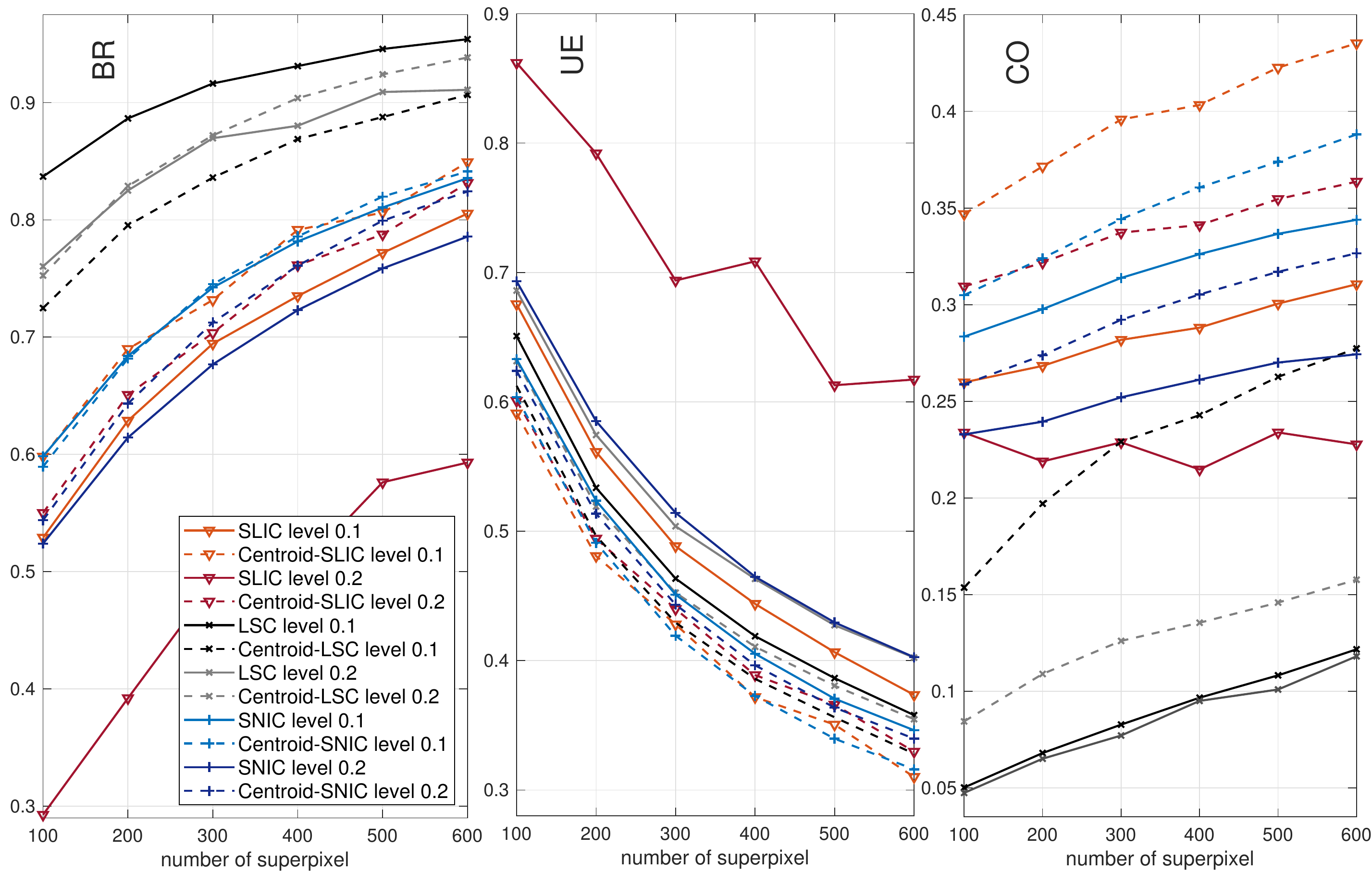}}
	\caption{The comparison of $X$ and Centroid-\emph{X} in Gaussian noise environment (level means the std of the noise).}
	\label{fig6}
\end{figure}

\begin{figure*}[htbp]
	\centerline{\includegraphics[width = 1.64\columnwidth]{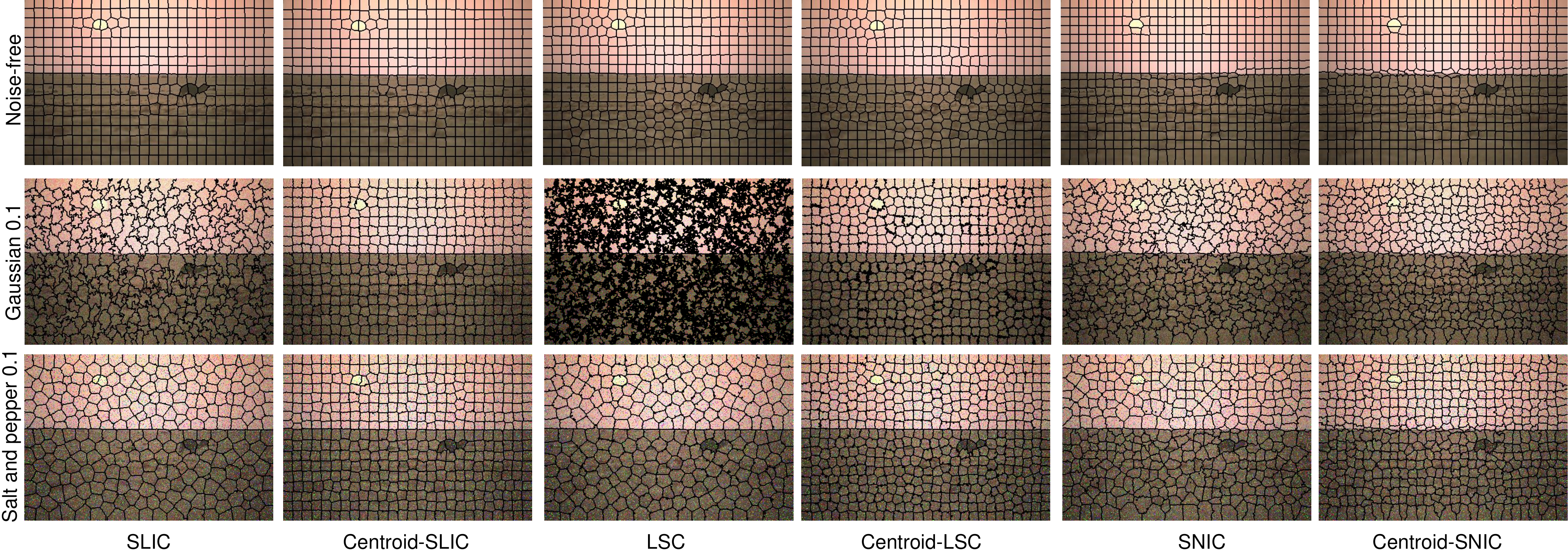}}
	\caption{The visual comparison of $X$ and Centroid-\emph{X} in noise-free environment and noisy environment (the desired number of superpixels for all methods is set as 600).}
	\label{fig5}
\end{figure*}

Fig. \ref{fig6} shows the comparison of $X$ and Centroid-\emph{X} in Gaussian noise environment. We can see that Centroid-SLIC and Centroid-SNIC obtain much better BR, UE and CO than SLIC and SNIC. Although LSC obtains better BR than Centroid-LSC, its UE and CO are significantly worse than Centroid-LSC. In Fig .\ref{fig5}, we can see that the performance of LSC in Gaussian noise environment are quite bad (like over-segmentation), so do SLIC and SNIC, while their corresponding Centroid-\emph{X} can still maintain the approximate performance like in noise-free environment.
\begin{figure}[htbp]
	\centerline{\includegraphics[width = 0.8\columnwidth]{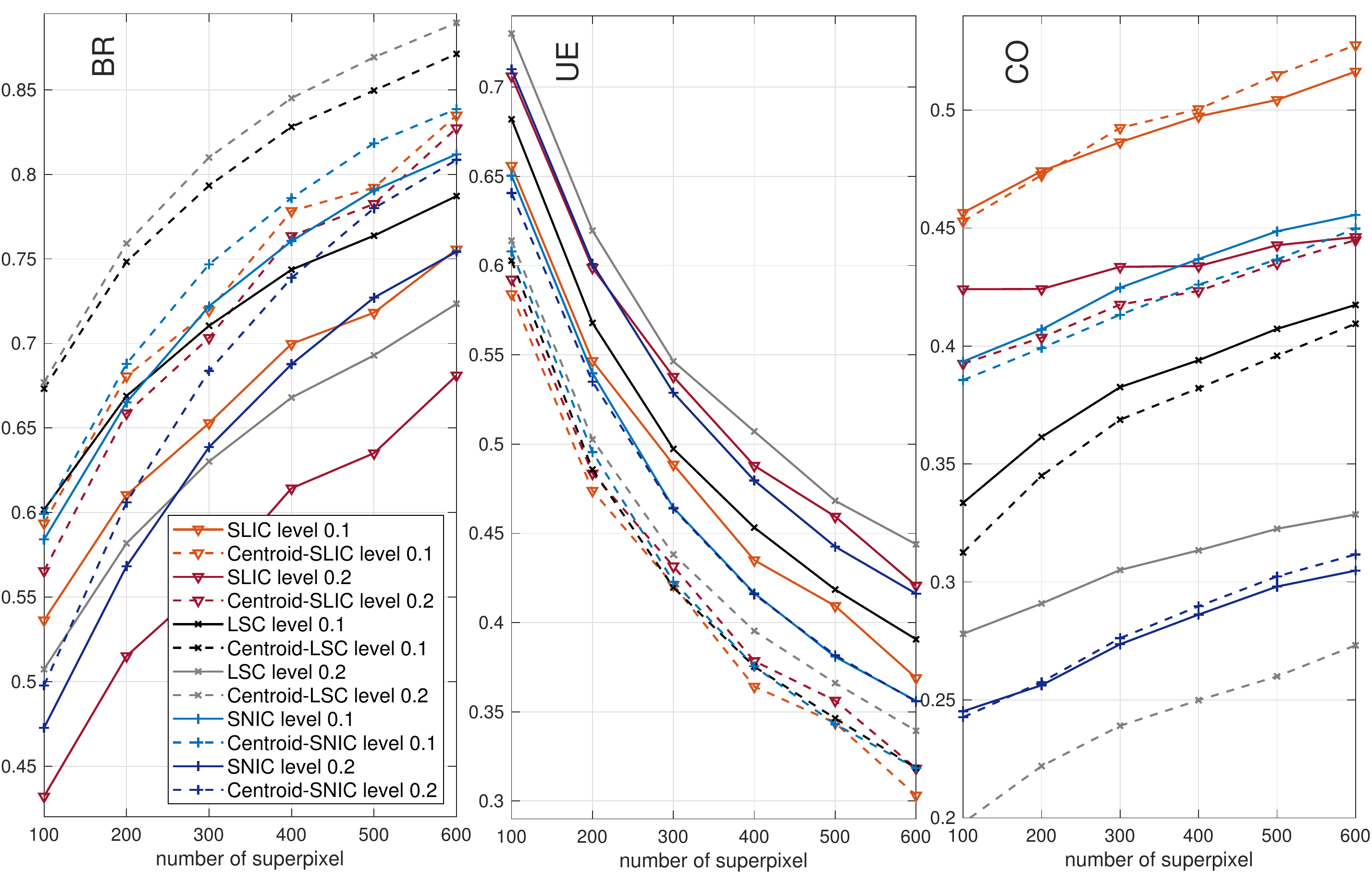}}
	\caption{The comparison of $X$ and Centroid-\emph{X} in salt and pepper noise environment (level means the density of the noise).}
	\label{fig7}
\end{figure}

Fig. \ref{fig7} shows the performance of $X$ and Centroid-\emph{X} in salt and pepper noise environment. By observing Fig. \ref{fig5}, we can find that the number of the output superpixels of SLIC and LSC falls more sharply (like under-segmentation) in salt and pepper noise environment than noise-free environment. But Centroid-SLIC and Centroid-LSC can still maintain comparable number of output superpixels. Although CO of Centroid-\emph{X} is slightly weaker than $X$, Centroid-\emph{X} still obtains much better BR and UE, which illustrates that even in salt and pepper noise environment, Centroid-\emph{X} can still get a robust performance like in noise-free environment.

\section{Application}
Here, we apply Centroid-SLIC into our SBED to generate superpixel, and we compare SBED with classical edge detection methods like Sobel \cite{kanopoulos1988design} and Canny \cite{canny1986computational}. We take PSNR \cite{hore2010image} and SSIM \cite{wang2004image} as evaluation metrics like \cite{he2019novel} on BSD500. Here we set $G_{low} = 0.1*\max(G)$, and $G_{high} = 0.8*\max(G)$. For Sobel and Canny, we set threshold as 0.1.
\begin{table}[h]
	\centering
	\caption{Comparison of three edge detection methods.} 
	\label{Tab1} 
	\scalebox{1}{
		\begin{tabular}{c c c c}
			\hline
			& Our method & Sobel & Canny\\ 
			\hline 
			PSNR & 6.8372 & 6.5926 & 6.1129  \\ 
			SSIM & 0.0117  & 0.0112 & 0.0104 \\    
			\hline
	\end{tabular}}
\end{table}

\begin{figure}[htbp]
	\centerline{\includegraphics[width = 0.9\columnwidth]{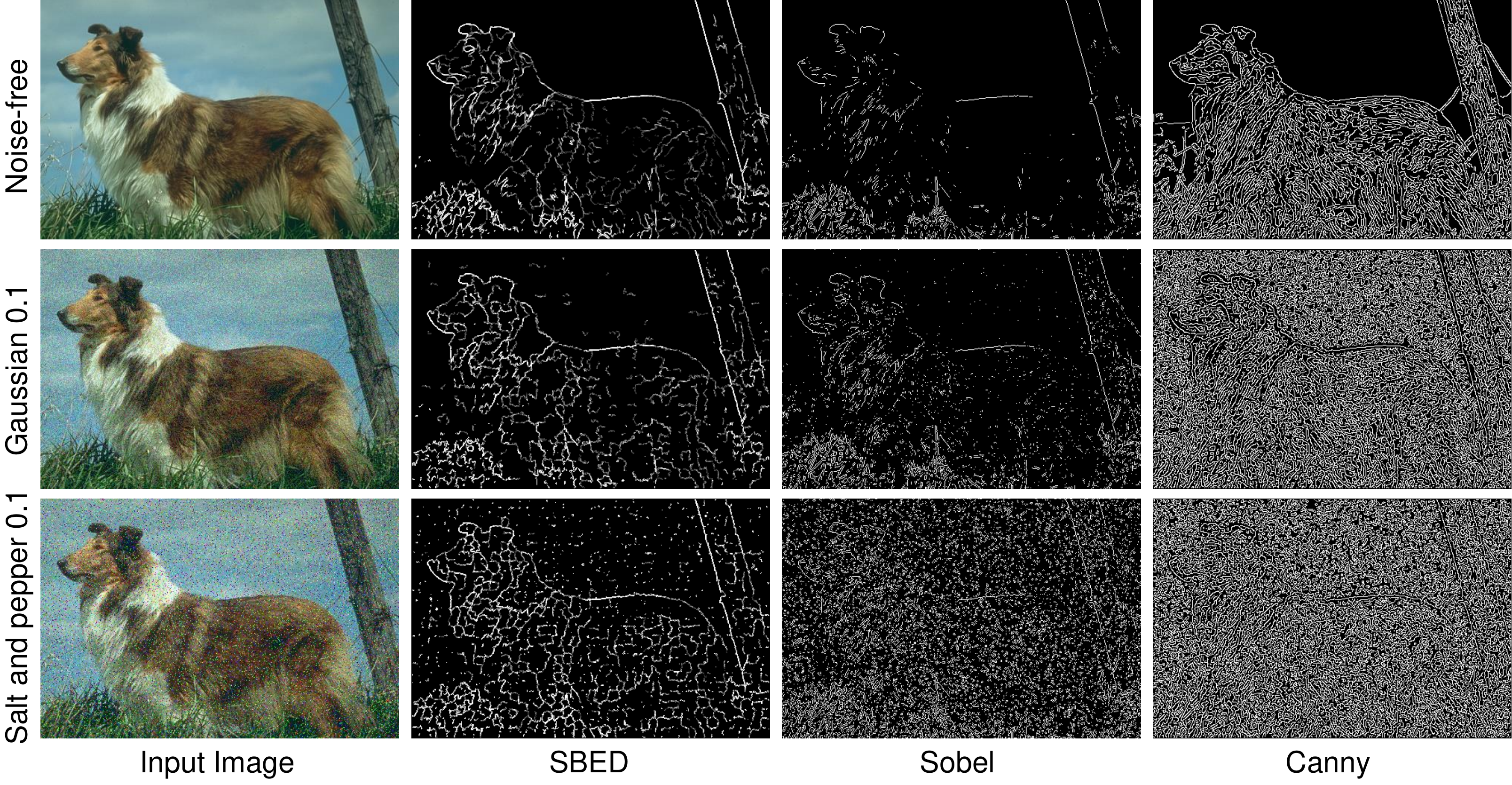}}
	\caption{The visual comparison of three edge detection methods in noise-free and noisy environment (in SBED, $k$ is set as 1500).}
	\label{fig8}
\end{figure}

Table \ref{Tab1} shows that our method gets better PSNR and SSIM than Sobel and Canny under noise-free environment. And Fig. \ref{fig8} shows that our method can better obtain the edge of image in both noise-free and noisy environment.

\section{Conclusion}
In this paper, we propose a novel centroid update approach to enhance clustering-based superpixel methods and a superpixel-based edge detection method. Experiments illustrate that our proposed methods can get a much better performance in noisy environment compared to state-of-the-art methods.

\section{Copyright}
\copyright \ 2020 IEEE. Personal use of this material is permitted. Permission from IEEE must be obtained for all other uses, in any current or future media, including reprinting/republishing this material for advertising or promotional purposes, creating new collective works, for resale or redistribution to servers or lists, or reuse of any copyrighted component of this work in other works.

\bibliographystyle{IEEEbib}
\bibliography{mybibfile}

\begin{thebibliography}{10}

\bibitem{ren2003learning}
Xiaofeng Ren and Jitendra Malik,
\newblock ``Learning a classification model for segmentation,''
\newblock in {\em Proceedings of the IEEE International Conference on Computer
  Vision}. IEEE, 2003, p.~10.

\bibitem{fslic}
C.~{Wu}, L.~{Zhang}, H.~{Zhang}, and H.~{Yan},
\newblock ``Superpixels using fuzzy simple linear iterative clustering and fast
  precise number control,''
\newblock {\em arXiv}, p. 1812.10932v2, 2018.

\bibitem{ICIP}
C.~{Wu}, L.~{Zhang}, H.~{Zhang}, and H.~{Yan},
\newblock ``Improved superpixel-based fast fuzzy c-means clustering for image
  segmentation,''
\newblock in {\em 2019 IEEE International Conference on Image Processing
  (ICIP)}, Sep. 2019, pp. 1455--1459.

\bibitem{accw}
C.~{Wu}, L.~{Zhang}, J.~{Cao}, and H.~{Yan},
\newblock ``Superpixel tensor pooling for visual tracking using multiple
  midlevel visual cues fusion,''
\newblock {\em IEEE Access}, pp. 1--1, 2019.

\bibitem{achanta2017superpixels}
Radhakrishna Achanta and Sabine Susstrunk,
\newblock ``Superpixels and polygons using simple non-iterative clustering,''
\newblock in {\em Proceedings of the IEEE Conference on Computer Vision and
  Pattern Recognition}, 2017, pp. 4651--4660.

\bibitem{achanta2012slic}
Radhakrishna Achanta, Appu Shaji, Kevin Smith, Aurelien Lucchi, Pascal Fua, and
  Sabine S{\"u}sstrunk,
\newblock ``Slic superpixels compared to state-of-the-art superpixel methods,''
\newblock {\em IEEE transactions on pattern analysis and machine intelligence},
  vol. 34, no. 11, pp. 2274--2282, 2012.

\bibitem{jampani2018superpixel}
Varun Jampani, Deqing Sun, Ming-Yu Liu, Ming-Hsuan Yang, and Jan Kautz,
\newblock ``Superpixel sampling networks,''
\newblock in {\em Proceedings of the European Conference on Computer Vision
  (ECCV)}, 2018, pp. 352--368.

\bibitem{guo2018fuzzy}
Yuwei Guo, Licheng Jiao, Shuang Wang, Shuo Wang, Fang Liu, and Wenqiang Hua,
\newblock ``Fuzzy superpixels for polarimetric sar images classification,''
\newblock {\em IEEE Transactions on Fuzzy Systems}, vol. 26, no. 5, pp.
  2846--2860, 2018.

\bibitem{stutz2018superpixels}
David Stutz, Alexander Hermans, and Bastian Leibe,
\newblock ``Superpixels: An evaluation of the state-of-the-art,''
\newblock {\em Computer Vision and Image Understanding}, vol. 166, pp. 1--27,
  2018.

\bibitem{li2015superpixel}
Zhengqin Li and Jiansheng Chen,
\newblock ``Superpixel segmentation using linear spectral clustering,''
\newblock in {\em Proceedings of the IEEE Conference on Computer Vision and
  Pattern Recognition}, 2015, pp. 1356--1363.

\bibitem{kslic}
Houwang Zhang and Yuan Zhu,
\newblock ``Kslic: K-mediods clustering based simple linear iterative
  clustering,''
\newblock in {\em Chinese Conference on Pattern Recognition and Computer Vision
  (PRCV)}. Springer, 2019, pp. 519--529.

\bibitem{buades2005non}
Antoni Buades, Bartomeu Coll, and J-M Morel,
\newblock ``A non-local algorithm for image denoising,''
\newblock in {\em 2005 IEEE Computer Society Conference on Computer Vision and
  Pattern Recognition (CVPR'05)}. IEEE, 2005, vol.~2, pp. 60--65.

\bibitem{zhang2014new}
Peixuan Zhang and Fang Li,
\newblock ``A new adaptive weighted mean filter for removing salt-and-pepper
  noise,''
\newblock {\em IEEE signal processing letters}, vol. 21, no. 10, pp.
  1280--1283, 2014.

\bibitem{kanopoulos1988design}
Nick Kanopoulos, Nagesh Vasanthavada, and Robert~L Baker,
\newblock ``Design of an image edge detection filter using the sobel
  operator,''
\newblock {\em IEEE Journal of solid-state circuits}, vol. 23, no. 2, pp.
  358--367, 1988.

\bibitem{arbelaez2010contour}
Pablo Arbelaez, Michael Maire, Charless Fowlkes, and Jitendra Malik,
\newblock ``Contour detection and hierarchical image segmentation,''
\newblock {\em IEEE transactions on pattern analysis and machine intelligence},
  vol. 33, no. 5, pp. 898--916, 2010.

\bibitem{schick2014evaluation}
Alexander Schick, Mika Fischer, and Rainer Stiefelhagen,
\newblock ``An evaluation of the compactness of superpixels,''
\newblock {\em Pattern Recognition Letters}, vol. 43, pp. 71--80, 2014.

\bibitem{levinshtein2009turbopixels}
Alex Levinshtein, Adrian Stere, Kiriakos~N Kutulakos, David~J Fleet, Sven~J
  Dickinson, and Kaleem Siddiqi,
\newblock ``Turbopixels: Fast superpixels using geometric flows,''
\newblock {\em IEEE transactions on pattern analysis and machine intelligence},
  vol. 31, no. 12, pp. 2290--2297, 2009.

\bibitem{veksler2010superpixels}
Olga Veksler, Yuri Boykov, and Paria Mehrani,
\newblock ``Superpixels and supervoxels in an energy optimization framework,''
\newblock in {\em European conference on Computer vision}. Springer, 2010, pp.
  211--224.

\bibitem{canny1986computational}
John Canny,
\newblock ``A computational approach to edge detection,''
\newblock {\em IEEE Transactions on pattern analysis and machine intelligence},
  , no. 6, pp. 679--698, 1986.

\bibitem{hore2010image}
Alain Hore and Djemel Ziou,
\newblock ``Image quality metrics: Psnr vs. ssim,''
\newblock in {\em 2010 20th International Conference on Pattern Recognition}.
  IEEE, 2010, pp. 2366--2369.

\bibitem{wang2004image}
Zhou Wang, Alan~C Bovik, Hamid~R Sheikh, Eero~P Simoncelli, et~al.,
\newblock ``Image quality assessment: from error visibility to structural
  similarity,''
\newblock {\em IEEE transactions on image processing}, vol. 13, no. 4, pp.
  600--612, 2004.

\bibitem{he2019novel}
Y~He and LM~Ni,
\newblock ``A novel scheme based on the diffusion to edge detection.,''
\newblock {\em IEEE transactions on image processing: a publication of the IEEE
  Signal Processing Society}, vol. 28, no. 4, pp. 1613--1624, 2019.

\end{thebibliography}

\end{document}